%% file: qualitycomp.tex
\documentclass[11pt,a4paper]{article}
\usepackage[hyperref]{emnlp-ijcnlp-2019}
\usepackage{times}
\usepackage{latexsym}
\usepackage{enumitem}
\usepackage{amsmath}
\usepackage{url}
\usepackage{multirow}
\usepackage{textcomp}
\usepackage{graphicx}

\aclfinalcopy 

\title{Comparison of Quality Indicators in User-generated \\Content Using Social Media and Scholarly Text}

\author{
    {\large Manirupa Das} \& 
    {\large Renhao Cui} \\
    {\large The Ohio State University} \\
    {\tt \{das.65, cui.182\}@osu.edu } \\
}

\begin{document}

\maketitle
\input{abstract}
\input{introduction}
\input{related_work}
\input{features}
\input{experiments}
\input{conclusion}

\bibliographystyle{apa}
\bibliography{qualitycomp}

\end{document}

%% file: abstract.tex
\begin{abstract}
Predicting the quality of a text document is a critical task when presented with the problem of measuring the performance of a document before its release. In this work, we evaluate various features including those extracted from the text content (textual) and those describing higher-level characteristics of the text (meta) features that are not directly available from the text, and show how these features inform prediction of document quality in different ways. Moreover, we also compare our methods on both social user-generated data such as tweets, and scholarly user-generated data such as academic articles, showing how the same features differently influence prediction of quality across these disparate domains.
\end{abstract}

%% file: introduction.tex
\section{Introduction}
Data is being produced  at an exponential rate online, with 90 percent of the data in the world generated over the last two years alone \citep{marr2018data}. This includes tweets, blogs, news, scholarly publications and a host of social content. While social media is more popular nowadays, standard text content is still an important method for disseminating information and ideas. 
Formal texts more amenable to syntactic parsing, such as news and academic articles, have been well--studied in terms of summarization, topic modeling, opinion mining, information extraction, etc. On the other hand, social media--based texts such as tweets, statuses, and text messages add more challenge to the classical task, especially when considering out-of-vocabulary words, length, and non-standard grammar. For many text analysis tasks, such as predicting the influence or factuality of articles or online content, predicting the quality of a text document becomes both useful and interesting. In this paper, we explore different factors that could affect the ``quality'' of a document, as well as provide a comparison between different types of unstructured data to investigate the role of each factor. We conduct experiments to study the effects of our chosen quality indicators on a dataset of 110k tweets for user-generated social data and on the ACL anthology of more than 20K documents until year 2014, for user-generated scholarly text data. Since the quality of a text document is a very subjective measurement, in this work, we use the Pagerank score as the class label for the scholarly articles dataset, similar to \cite{yogatama2011predicting}, and the number of retweets as the indicator of quality for the tweets dataset, similar to \cite{hong2011predicting}. Our experiments reveal several interesting and surprising findings regarding the influence that various textual and meta-level features have on the quality of a document, based on its domain. 

%% file: related_work.tex
\section{Related Works}
While numerous works exist in the literature for prediction, analysis and assessment of document quality \cite{hong2011predicting, yogatama2011predicting, dalip2011automatic, coleman1975computer, de2012did, egbert2015success}, to the best of our knowledge we are the first to perform such a comparative study of similar quality indicators across disparate domains of user-generated text. We cast the problem of predicting a document\textquotesingle s quality, or importance, as a task in supervised learning, using an available signal from the data as a proxy for the notion of quality, and apply several methods towards feature analysis and actual training of our datasets. In the sections below, we therefore describe in more detail each of the related areas pertaining to the methods employed to obtain these features, that are used as the indicators of quality in our experiments.

\subsection{Topic Modeling}
Topic modeling assumes that each word in a collection carries some meaning towards a set of topics, where each document, treated as a combination of the words, then carries some meaning with respect to the set of topics in the collection. Thus a topic model learns a probability distribution for a set of topics across the set of documents in a collection. We employ Latent Dirichlet Allocation (LDA)\cite{blei2003latent} to obtain topic distributions for each document in our datasets. We infer 30 topics for our models obtaining a 30-dimensional real-valued feature vector for each document.

\subsection{Part-of-speech Analysis}
Part-of-speech tagging is used to distinguish the role of the word within a sentence, and to better understand the structure of the sentence by annotating each word with a part-of-speech marker. 
Some words have multiple part-of-speech tags, so context is taken into account to break ties to obtain the POS tag. A chain model that carries the information from the previous predictions during tagging, is used to obtain the context information. We consider all tags available in the Penn Treebank and obtain a 38-dimensional binary-valued feature vector per document in this way.

\subsection{Readability Analysis}
Readability scores measure how easy it is to read and understand a document. It tries to predict the reaction from humans as they read the document. In this work, we consider several readability metrics. The Coleman–Liau Index of readability measurement \cite{coleman1975computer, dalip2011automatic} is calculated as: 
\begin{align*}
CLI = 0.0588 \times L - 0.296\times S - 15.9
\end{align*}
where $L$ is the average number of letters per 100 words and $S$ is the average number of sentences per 100 words. 

The ARI metric \cite{dalip2011automatic} where the authors analyzed samples of the textbooks used in the Cincinnati Public School System from which they derived the multiple regression formula given by:  
\begin{align*}
ARI = 4.71 \times \frac{characters}{words} + \\0.5 \times \frac{words}{sentence} - 21.43
\end{align*}
It uses the average number of words per sentence and the average number of characters per word to estimate the readability. 

The GFI metric \cite{dalip2011automatic} is given by: 
\begin{align*}
GFI = 0.4 \times (\frac{words}{sentences} + \\100 \times \frac{complexwords}{words})
\end{align*}
The score uses the average number of words per sentence and the average number of complex words in the text, such that short sentences in plain English achieve a better score than long sentences written in a complicated language. 
A complex word is defined as a word with three or more syllables. Each of these metrics gives us a real-valued feature.

\subsection{Sentiment Analysis}
Sentiment analysis is used to extract subject information from text content, and determine the attitude of the author with respect to the overall polarity of the text. 
Generally this type of analysis is performed on a word-to-word or sentence basis, and then combined together as the sentiment of a document. 
Various models have been applied to this problem, and most of them are treated as a classification problem, with a deterministic or probabilistic output indicating the polarity. 

\section{Datasets}
\subsection{Tweets}
We extract English tweets through Twitter streaming API across 7 days, and result in about 3.5 million tweets. 
Since the task is to predict the influence of a tweet, we only pick the original tweets, then filter out duplicates and strange-formatted tweets that are not easy to process. 
We use the number of retweets as the class label, and format a binary decision similar to \cite{hong2011predicting}: whether this tweet got any retweets (class 2) or not (class 1) . 
Finally we trim the dataset to balance the data for both labels, and it results in 110k tweets for this work.
 
\subsection{Articles}
The ACL Anthology \cite{bird2008acl} is a digital archive of conference and journal papers in natural language processing and computational linguistics whose primary purpose is to serve as a reference repository of research results, but the creators believe that it can also be an object of study and a platform for research in its own right and have made this dataset a public resource that can be used for research in scholarly document processing. 
We use the 2013 release of this dataset, spanning 21,212 papers, 17,792 authors, 342 venues and 110,975 citations.
 
The Pagerank statistic available from the dataset, was used as the class label for articles since it measures the importance of an article within a network. Here we looked at the real values of this statistic and choose the threshold as the lowest positive value of Pagerank that corresponds to 4163 articles out of the 11592 articles, to convert this to a categorical value giving a binary decision: well-cited paper (Y) versus not well-cited paper (A). 
We trim the dataset to balance both labels, and since only papers with positive values of Pagerank were chosen every paper considered has definitely been cited, hence the binary decision is for moderate versus high importance as opposed to cited yes/no, and it results in a dataset of 11592 articles for this work.

%% file: features.tex
\section{Methodology}

\subsection{Logistic Regression}
Conditional or discriminative probabilistic models such as maximum entropy and logistic regression have been extensively used in NLP, IR, and Speech. They give good performance and make it easy to incorporate lots of linguistically important features.  They also allow automatic building of language independent and retargetable NLP modules \cite{manning2003optimization}. These discriminative (conditional) models take the data as given, and put a probability over hidden structure given the data. In this work, we therefore use logistic regression as the predictive model for our experiments as it is useful in estimating the parameters of a qualitative response model \cite{kleinbaum2002logistic}.
 
The model can be used for predicting the outcome of a categorical dependent variable, based on one or more explanatory variables; and it can also be used for regression to predict a real-valued number, which can also be thresholded to obtain a binary classification. 
The probabilities describing the possible outcomes of a single trial are modeled as a function of the explanatory (predictor) variables or features, using a logistic function, as shown. 
\begin{align*}
F(t) = \frac{e^t}{e^t+1} = \frac{1}{1+e^{-t}}
\end{align*}                                                           
Since the Logistic function squashes linear response into a real value between 0 and 1, it is interpretable as a probability \cite{kleinbaum2002logistic}, and therefore, for regression, we can use the output directly, whereas for binary classification we can label the output as the class with the higher probability. Thus this model is perfectly suited to our task of predicting PageRank or Retweet--based class labels.

\subsection{Feature Extraction}

\subsubsection{Meta Features}
The Meta feature is used to describe some high level characteristics other than textual content. 
In this work, for tweets, we use the number of followers, the number of favorites, and the number of posted tweets of the author as meta features. 
These can represent the popularity of the author, which related to how many people can see the posts, and the influence of the author.
Since we use the number of retweets as the label, these features can also be used to balance the activity level of the author. 

For the scholarly article dataset, the meta feature set includes fields such as author in-citations, author out-citations, author Pagerank, publication year, and publication venue, and it also includes some network-based features such as paper between-ness centrality, closeness centrality, degree centrality, paper in-citations, and paper out-citations.
 
\subsubsection{Textual Features}
Text features are those extracted from the textual content of the articles or tweets \cite{dalip2011automatic}. 
We extract two types of textual features from the documents, a document coherence feature and a part-of-speech feature. 
Apart from this we also include length features such as word counts, character counts and sentence counts to represent the text.
\begin{itemize}
    \item  \textbf{Coherence/Topic Feature} In general, a document focuses on several key ideas that are used to express the opinion of the author. We believe the way the author organizes these ideas is related to the quality of the document. In this case it will affect the influence of the document. To capture the arrangement of these ideas, we use topic modeling to generate the distribution of topics for a document. Thus, we treat this probability distribution as representative of the arrangement of the ideas. Since most documents focus on a small number of topics, and to prevent noisy information as well as help the performance, we decide to take the top-5 topics probabilities for a document as the coherence feature by default, although we perform the topic modeling for 30 topics. For this work, we use Stanford Topic Modeling Toolbox \cite{ramage2011stanford} for LDA--based topic modeling. 
    Thus we have two sets of textual ``coherence'' features, the set of 30 topic distributions, and the set of top-5 normalized topic distributions for each document.

    \item \textbf{POS Feature} POS tags are used to capture the role and type of  words within the sentence. However, structured or formal and unstructured or informal documents have very different writing style, so we decide to use domain specific POS tags for this task. For articles, we use the standard Penn Treebank POS tag set available within the NLTK package \cite{loper2002nltk} used for generating POS statistics for the article corpus with a total of 38 tags. For tweets, we used the twitter-based POS tagger from O’Connor to generate a total of 24 tags \cite{owoputi2012part}.

    \item \textbf{Length Feature} Document length varies from one to the other, and it is closely related to whether the reader would prefer the content. It is also critical to the POS feature, as we do not normalize the feature, thus we put the length as one of the features instead. 
    This includes the sentence counts, word counts and character counts per document in case of articles, and the word token counts in case of tweets.
    
\end{itemize}

\subsubsection{Readability Feature}
It is not necessarily true that an easily understandable document has more influence among the readers. However, we think there is some relation between these two, so we use the Coleman–Liau index as the readability feature. For the scholarly articles we also calculate and include the Automated Readability Index and Gunning-Fog Index readability measures described earlier as we want to take the role of average words/sentence, average characters/word and complex words into account for longer structured text such as articles.

\subsubsection{Sentiment Feature} 
Sentiment analysis is used to analyze the opinion that the text expresses to be positive or negative. To be more specific, a binary indication of the sentiment is too general for this task, so we use a confidence probability score to represent the sentiment of the text. Therefore it gives more quantitative measurement of the sentiment of the document. For the tweet dataset, we use a sentiment word list with probabilities created using a Language-independent Bayesian method from Davies and Ghahramani \cite{davies2011language}. 

In the case of articles, we used the TextBlob API \cite{loria2014textblob} for processing of the text for extracting the sentiment feature, and we consider two measures of sentiment, the average sentiment over all sentences in the article and also the maximum sentiment or maximum value of polarity for that article.


%% file: experiments.tex
\section{Experiments}

Based on the individual features mentioned in the previous section, we separate them into 4 categories:
\begin{enumerate}[label=(\roman*)]
\item Metadata/Network Feature -- features that are not related to the actual text content of the document, and used to describe some background of the author and the document (tweet or article) and their peer network. In this work, this includes the Meta feature.
\item Textual Feature -- features that are directly related to the text content of the document. This includes the Coherence/Topic Feature, POS feature, and Length features.
\item Readability Feature -- this feature represents the degree to which a reader can read and understand the document. In this work, this includes the different readability metrics.
\item Sentiment Feature -- this feature represents the opinion--based measures for the document. In this work, these are the sentiment features.
\end{enumerate}

In order to explore the role of different features as well as their categories, we run experiments on the 16 different models with various feature sets as shown in Table 1. 
Here Models 1-4 represent models having each individual category of features, Model 5 has all feature sets, hence represent the ``Full'' model using top-5 topic features. Model 6 is a Full model with all 30 topics. Models 7-14 represent feature set ablations from the Full model, while Models 15 \& 16 represent interesting feature set combinations.
\begin{table*} 
\caption{Model Feature sets and Ablations}
\centering
\begin{tabular}{c|l|c|l}
\hline
\textbf{No.}&\textbf{Description}&\textbf{No.}&\textbf{Description}\\
\hline\hline
\multirow{2}{*}{1}&\multirow{2}{*}{Meta/Network Feature}&\multirow{2}{*}{9}&Full Model without Coherence/Topic\\
&&&Feature\\
\hline
2&Readability Feature&10&Full Model without POS feature\\
\hline
3&Sentiment Feature&11&Full Model without Length Feature\\
\hline
\multirow{2}{*}{4}&Textual Feature (Coherence/Topic Feature&\multirow{2}{*}{12}&\multirow{2}{*}{Full Model without Meta Feature}\\
& + POS Features + Length Feature)&&\\
\hline
\multirow{2}{*}{5}&Full Model (Meta Feature + Readability&\multirow{2}{*}{13}&\multirow{2}{*}{Coherence/Topic Feature with 5 topics}\\
&Feature + Sentiment Feature + Textual Feature)&&\\
\hline
\multirow{2}{*}{6}&Full Model with 30 topics in &\multirow{2}{*}{14}&\multirow{2}{*}{Coherence/Topic Feature with 30 topics}\\
&Coherence/Topic Feature&&\\
\hline
7&Full Model without Readability Feature&15&Textual Feature + Sentiment Feature\\
\hline
8&Full Model without Sentiment Feature&16&Meta Feature + Sentiment Feature\\
\hline
\end{tabular}
\end{table*}

We use the Weka data mining toolkit \cite{hall2009weka} to implement the logistic regression classifier, and use its default settings. 
We apply 10-fold cross validation for our training and testing dataset split. 
We record results for metrics such as Precision, Recall, and F1-score for each class and the weighted average for each, and also the overall accuracy of the model.

\section{Results}

\subsection{Tweets Dataset}
The models are evaluated on the 110k tweet dataset, and we record the Accuracy, F1 scores for Class 1 (no retweet) and Class 2 (got retweeted), Average F1 score, Average Precision and Recall. 
The data is all labeled with 54.6\% in Class 1.



\begin{table*}
\caption{Model Performance on \textbf{Tweets} Dataset}
\centering
\begin{tabular}{l|c|c|c|c|c|c}
\hline
\textbf{Features}&\textbf{Accuracy}&\textbf{F1 Class 1}&\textbf{F1 Class 2}&\textbf{F1}&\textbf{Precision}&\textbf{Recall}\\
\hline\hline
\textbf{Meta/Network}&58.05&71.1&23.5&49.5&62.1&58.0\\
\hline
\textbf{Textual (Topics + POS}&\multirow{2}{*}{56.97}&\multirow{2}{*}{64.2}&\multirow{2}{*}{46.1}&\multirow{2}{*}{56.0}&\multirow{2}{*}{56.4}&\multirow{2}{*}{57.0}\\
\textbf{+ Length)}&&&&&&\\
\hline
\textbf{Readability}&54.68&\textbf{70.2}&\textbf{5.4}&40.8&53.3&54.7\\
\hline
\textbf{Sentiment}&54.60&\textbf{70.6}&\textbf{0.0}&38.6&29.8&54.6\\
\hline
\textbf{Full 5 topics}&\textbf{60.66}&\textbf{67.7}&\textbf{49.7}&59.5&60.4&60.7\\
\hline
\textbf{Full 30 topics}&\textbf{60.69}&67.6&50.0&59.6&60.4&60.7\\
\hline
\textbf{Full no sentiment}&60.14&67.5&48.5&58.9&59.9&60.1\\
\hline
\textbf{Full no readability}&60.63&67.7&49.6&59.5&60.4&60.6\\
\hline
\textbf{Full no topics}&\textbf{60.90}&68&49.7&59.7&60.7&60.9\\
\hline
\textbf{Full no POS}&58.57&69.5&35.5&54.1&59.2&58.6\\
\hline
\textbf{Full no length}&60.49&67.6&49.5&59.4&60.2&60.5\\
\hline
\textbf{Full no network/meta}&57.47&64.3&47.5&56.6&56.9&57.5\\
\hline
\textbf{Coherence only (5 topics)}&54.36&69.9&6.1&40.9&50.8&54.4\\
\hline
\textbf{Coherence only (30 topics)}&54.23&69.5&8.8&41.9&50.8&54.2\\
\hline
\textbf{Meta/Network + Sentiment}&58.11&71.1&23.9&49.7&62.1&58.1\\
\hline
\textbf{Textual (Topics + POS }&\multirow{2}{*}{57.57}&\multirow{2}{*}{64.4}&\multirow{2}{*}{47.4}&\multirow{2}{*}{56.7}&\multirow{2}{*}{57.0}&\multirow{2}{*}{57.6}\\
\textbf{+ Length) + Sentiment}&&&&&&\\
\hline
\end{tabular}
\end{table*}

Table 2 lists the overall performance of the models on Tweet dataset.
It can be seen that full models give relatively good performance. However, excluding coherence/topic feature outperforms the full model.
Class 1 has a much better F1-scores compared to Class 2. It shows that the system works better in predicting a tweet to have no retweets.


In general, no single set of features outperforms the full model. 
The Meta feature has the best result among them, and it even gives a better result than the textual feature. 
It indicates that not knowing anything about the content, but knowing the author is most predictive of the quality of the tweet. 
This is actually similar to some previous work that claims meta or follower retweet counts have a huge impact on the influence of the tweet \cite{khabiri2009analyzing, petrovic2011rt}. 
Moreover, the sentiment or readability feature does not carry much semantic meaning, so it cannot be used as a good single predictor for tweets. 
We also notice that more topics as the summary of the content does not have a clear influence on the prediction for tweets.

Considering the feature ablation studies, removing the POS or Meta feature sets result in the most decrease in performance. 
It further proves that the Meta feature has a huge impact on the overall prediction. 
Meanwhile, POS features, related to the textual content, also has a considerable impact, which indicates that the syntactical use of words and maintaining grammatical structure of the sentence may bring gains in perceived quality of the tweet. 
On the other hand, removing readability and length features do not have much impact on the system performance, indicating the quality of tweets is not mainly judged by the complexity of the writing.
Furthermore, removing the sentiment feature results in some decrease, which agrees with the previous discussion that sentiment does not inform much to the prediction of tweet quality. 
But it has more positive impact on a model that has relatively poor performance (e.g. Metadata vs. Textual).

Finally, removing topic features improves the performance. 
The use of topics to summarize the text content may not help in predicting the quality, and it also suggests that non-semantic features contribute more to the prediction. 
The worst performance of using coherence/topic feature alone also shows the weakness of the semantic-related features.

To sum up, using all features gives a relatively good performance. However, metadata only also has a huge impact on the prediction. Sentiment seems to be an average indicator for this task on tweets, while readability and length features are not the big predictors either. Moreover, the use of topics as a summary does not play a good role as hypothesized, and even has a negative impact in some cases. The choice of number of topics does not clearly make a difference either. 

\begin{table*}
\caption{Model Performance on the \textbf{Articles} Dataset}
\centering
\begin{tabular}{l|c|c|c|c|c|c}
\hline
\textbf{Features}&\textbf{Accuracy}&\textbf{F1 Class A}&\textbf{F1 Class Y}&\textbf{F1}&\textbf{Precision}&\textbf{Recall}\\
\hline\hline
\textbf{Meta/Network}&\textbf{86.28}&81.4&89.1&86.3&86.5&86.3\\
\hline
\textbf{Textual (Topics + POS}&\multirow{2}{*}{67.96}&\multirow{2}{*}{26.0}&\multirow{2}{*}{79.6}&\multirow{2}{*}{60.3}&\multirow{2}{*}{70.5}&\multirow{2}{*}{68.0}\\
\textbf{+ Length)}&&&&&&\\
\hline
\textbf{Readability}&65.57&27.9&77.4&59.6&63.0&65.6\\
\hline
\textbf{Sentiment}&67.69&25.2&79.4&59.9&69.9&67.7\\
\hline
\textbf{Full 5 topics}&86.74&81.8&\textbf{89.6}&86.8&86.8&86.7\\
\hline
\textbf{Full 30 topics}&\textbf{86.81}&81.9&\textbf{89.6}&86.9&86.8&86.9\\
\hline
\textbf{Full no sentiment}&86.65&81.7&89.5&86.7&86.8&86.6\\
\hline
\textbf{Full no readability}&\textbf{86.81}&81.9&89.6&86.9&86.9&86.8\\
\hline
\textbf{Full no topics}&86.49&81.5&89.4&86.5&86.6&86.5\\
\hline
\textbf{Full no POS}&86.75&82.0&89.5&86.8&87.0&86.7\\
\hline
\textbf{Full no length}&\textbf{86.77}&81.9&89.6&86.8&86.9&86.8\\
\hline
\textbf{Full no network/meta}&67.52&27.6&79.1&60.6&68.0&67.5\\
\hline
\textbf{Coherence only (5 topics)}&66.98&25.5&78.8&59.7&67.0&67.0\\
\hline
\textbf{Coherence only (30 topics)}&67.47&34.9&78.3&62.7&66.1&67.5\\
\hline
\textbf{Meta/Network + Sentiment}&\textbf{86.53}&81.8&89.3&86.7&86.8&86.5\\
\hline
\textbf{Textual (Topics + POS}&\multirow{2}{*}{68.05}&\multirow{2}{*}{26.4}&\multirow{2}{*}{79.6}&\multirow{2}{*}{60.5}&\multirow{2}{*}{70.6}&\multirow{2}{*}{68.0}\\
\textbf{+ Length) + Sentiment}&&&&&&\\
\hline
\end{tabular}
\end{table*}


\subsection{Articles Dataset}
We run the models on the dataset of 11592 articles with balanced Pagerank score, and record the performance metrics such as Accuracy, F1 scores for each class, Average F1 score, Average Precision and Recall. In balancing the dataset our default or majority class turns out to be class Y, i.e. for the well-cited papers, with 64.08\% of the dataset residing in this class. 
We perform most of our comparisons against this baseline.

Table 3 shows the model performance on the Articles dataset.
In the overall comparison we find, not too surprisingly, that the Full model with 30 topics as the coherence feature gives the best overall performance at 86.81\%. The Full model using top-5 topics comes in a close second at 86.74\% on accuracy, and nearly the same other scores. Hence we pick this model as the baseline for comparisons in all of our feature ablation experiments.
As in the case of articles, the majority class Y of well-cited papers has a much better F1 compared with class A of moderate importance papers, as can be expected. 
We know that ACL conference papers with Pagerank beyond our minimum positive threshold are likely to be well-cited within the community, i.e. at least better cited than the least cited papers within the community (i.e. class A), which our system correctly tends to predict.

In comparing with the individual feature models, no single model is able to surpass the performance of the full models with both 5 and 30 coherence features, though Metadata comes closest, at 86.28\%. 
Readability, Sentiment and Textual individual feature set models are the least predictive of quality
, though F1-scores show that they do predict the well-cited class correctly with fairly high confidence. Comparing all the scores for just the two Coherence features we find that the model with 30 topics does better overall than that with top 5 normalized topics.

Next, considering the feature ablations against the full model (here we use the full model with 5 topics for coherence, at 86.74\%), complementary to findings for individual models, we find that removing the Textual-based features such as Readability, POS and Length actually increases the performance of the system to 86.81\%, 86.75\% and 86.77\% respectively showing that these features are actually hurting prediction performance for Articles, whereas taking away the coherence features reduces performance indicating that these do contribute to the overall performance of the full model for Articles. As for the role of Sentiment, we see that when ``added'' to the individual models of Metadata, Textual and the Full model, it significantly contributes to the performance all-round, for Accuracy, Avg. F1, Avg. Precision and Avg. Recall. 

In this set of experiments we run the individual models for each feature set and compare the performance of each against the full model. Here we find that full model with the 30 topic features performs the best, with the non-textual metadata model performing the best among all the individual feature sets, with 86.28\% accuracy. Perhaps surprisingly, Sentiment and Readability models give the least performance, and do almost similarly with 67.69\% and 65.57\% accuracy respectively, and the Textual model does only slightly better than Sentiment and Readability with an accuracy of 67.96\% and precision at 70.5\% comparable to Sentiment at 69.9\%.

As seen previously, the removal of Metadata or Non-Textual features has the most negative impact on performance on the Full model, removing Coherence features (top-5 topics) also hurts performance, whereas removal of other textually based features like Readability, POS and Length actually increases performance indicating that these contribute little in terms of predictive value. 
The Full model also takes a hit on removal of sentiment, which hurts the performance.

As we can see from the results and previous discussion, Sentiment adds predictive value on all measures to each of the models, Metadata (Non-textual), Textual (Coherence+POS+Length) and Full models (top-5 topics). This highlights the fact that sentiment can play an important role in the how a scholarly article is received.

Thus, in the case of articles, \textbf{Metadata} features such as author, in-cites, out-cites, centrality measures, and Coherence/Topic features contribute the most predictive value to the overall model, whereas other textually derived features such as Readability, Length and POS can actually hurt the performance. 
Contrary to the tweets domain, sentiment does contribute predictive value to the overall model for scholarly articles.

%% file: conclusion.tex
\section{Conclusion and Future Work}
Besides the in-domain analysis of indicators of document quality, there are also some interesting differences and similarities in feature informativeness across domains. We find that for the tweets domain, sentiment does not have a noticeable influence on predicting whether a post will be retweeted or not. However, for articles, sentiment does bring an improvement over Metadata, Textual and Full models (5 topics). 
This seems to match with the intuition that tweets are primarily an opinion-based form of user-generated text where it does not add much more predictive value for tweet quality, whereas it does add a lot more predictive value to the quality of scholarly text. 
We also find that for both tweet and article domains Metadata or Non-Textual features play a significant role in increasing prediction accuracy, followed by Coherence/Topic features. 
This appeals to the intuition that this may be because people tend to relate the quality of a document with its author, or some other background factors. While our datasets are fairly balanced, this effect may also be due to the use of class labels that are inherently more biased towards the non-textual elements in both the datasets. 

Hence a different labeling scheme, perhaps manual annotation, could be used to see if this offsets the role of Metadata and boosts the role of Textual-based features. Since our work is the first effort known to us to perform a comprehensive analysis of quality indicators across very unrelated domains of user-generated text, and since our study constitutes an exploration of the most predictive feature sets for this task, 
we hope that our work will represent the baseline for such other comparisons across numerous interesting domains or sub-domains within social media text and scholarly text.

For future extensions to this work we would like to include context features into our models, e.g., referring texts such as new textual content added into a retweet for a new tweet, and explore the role of citing sentences \cite{abu2012reference} for articles. 
Additionally we would like to look at methods to improve the informativeness of our textual features as quality indicators, by possibly deriving semantic features by vocabulary expansion using WordNet and other ontologies, or via the use of skip gram features or top-ranking neighbor word embeddings \citep{mikolov2013distributed}. Since our study is mainly about trying to understand which features from the text are most indicative of the quality or importance of a document, we would also like to experiment with labels different from our current choices or a combination of labels, as there might be elements in our top performing Metadata features, that are biased towards predicting our particular choice of labels for both domains. 